\documentclass[runningheads]{llncs}
\usepackage{todonotes}
\usepackage{amsmath}
\usepackage{amssymb}
\usepackage{bbm}
\usepackage{adjustbox}
\usepackage{array,colortbl,multirow,multicol,booktabs,ctable}

\usepackage[title]{appendix}
\usepackage{xcolor}
\usepackage{setspace}

\usepackage[title]{appendix}
\usepackage{pbox}
\usepackage[most]{tcolorbox}
\usepackage{graphicx}
\graphicspath{{images/}}

\usepackage{threeparttable}
\usepackage[inline,shortlabels]{enumitem}
\usepackage{hyperref}
\hypersetup{
  colorlinks   = true, % Colors links instead of ugly boxes
  urlcolor     = blue, % Color for external hyperlinks
  linkcolor    = blue, % Color of internal links
  citecolor    = blue  % Color of citations
}
\usepackage{url}
\usepackage[T1]{fontenc}
\makeatletter

\makeatletter
\def\Url@twoslashes{\mathchar`\/\@ifnextchar/{\kern-.2em}{}}
\g@addto@macro\UrlSpecials{\do\/{\Url@twoslashes}}
\makeatother

\newcommand{\secref}[1]{\S\ref{#1}}
\newcommand{\appref}[1]{Appendix~\ref{#1}}

\newcommand{\figref}[1]{Fig.~\ref{#1}}

\newcommand{\tabref}[1]{Table~\ref{#1}}

\usepackage{color}

\begin{document}

\title{\textsc{SkillMatch}: Evaluating Self-supervised Learning of Skill Relatedness}
\titlerunning{\textsc{SkillMatch} Evaluation}

\author{Jens-Joris~Decorte\inst{1,2} \and
Jeroen~Van~Hautte\inst{2} \and
\\ Thomas~Demeester\inst{1}
\and
Chris~Develder\inst{1}
}

\authorrunning{J.-J.~Decorte \emph{et al.}}

\institute{Ghent University -- imec, 9052 Gent, Belgium \\
\urlstyle{rm}
\email{\{jensjoris.decorte, thomas.demeester, chris.develder\}@ugent.be}\\
\url{https://ugentt2k.github.io/}
\and
TechWolf, 9000 Gent, Belgium\\
\email{\{jensjoris, jeroen\}@techwolf.ai} \\
\url{https://techwolf.ai/}}

\maketitle

\begin{abstract}
Accurately modeling the relationships between skills is a crucial part of human resources processes such as recruitment and employee development.
Yet, no benchmarks exist to evaluate such methods directly.
We construct and release \textsc{SkillMatch}, a benchmark for the task of skill relatedness, based on expert knowledge mining from millions of job ads.
Additionally, we propose a scalable self-supervised learning technique to adapt a Sentence-BERT model based on skill co-occurrence in job ads.
This new method greatly surpasses traditional models for skill relatedness as measured on \textsc{SkillMatch}.
By releasing \textsc{SkillMatch} publicly, we aim to contribute a foundation for research towards increased accuracy and transparency of skill-based recommendation systems.
\end{abstract}

\section{Motivation and Related Work}
Skills play a central role in human resources (HR) processes such as recruitment or learning and development. 
Recommendation systems are often used in these processes to suggest actions based on skills.
Examples are ranking candidates for a job, or suggesting learning content to employees. 
However, relying solely on the explicit presence of a given skill would neglect the complex relations between skills in the real world.
In recruitment, taking related skills into account will increase the relevant candidate pool and ensure that all relevant candidates are considered.
Similarly, considering these related skills helps to more accurately identify skill gaps within an organization and effectively recommend appropriate courses to employees.

Several works study the usage of these skill relations as part of a job recommendation.
Static word vector methods have been used to compare resume and job ads content, including skills, in the context of job recommendation~\cite{jobrecskillemb,jobrecinproceedings}.
More recent approaches use a joint representation learning framework to represent jobs and skills based on graphs of job and skill relations~\cite{combinedreplearn,liu2019tripartite}.
In these works, evaluation solely focuses on downstream task performance.
However, it is unclear whether the obtained skill representations accurately reflect the relatedness between skills, given the absence of direct, intrinsic evaluation thereof.
Furthermore, evaluation on downstream tasks like job recommendation can hide task-specific biases in these representations, and does not ensure their validity for other uses.

Preliminary work on the intrinsic evaluation of skill representation methods relies on manual annotation by domain experts, and is therefore limited in scale.
An early method relies on semantic role features extracted from parse trees to capture skill similarity~\cite{pan-farrell-2007-computing}.
However, the corresponding human evaluation data, which comprises a limited set of 75 skill pairs with similarity rated on a five point scale, is not publicly available.
Similarly, Gandhi et al.~\cite{gandhi2022learning} use a set of 600 pairs of skills annotated on a three point scale for the evaluation of their method.
In another instance, qualitative feedback from subject matter experts is used to evaluate a learned model of skill relatedness in a large IT organization~\cite{vasudevan2018estimating}.
Le et al.~\cite{Le2017Skill2vecML} propose Skill2Vec, a method based on the popular Word2Vec method that learns static vectors for a vocabulary of skills, based on their presence in job ads.
They ask domain experts to manually select the relevant skills out of the five nearest neighbors, for 200 different samples.
For all the aforementioned works, the annotation effort required for their intrinsic evaluation is not scalable, and their data is not released.

With our work, we aim to tackle this lack of intrinsic evaluation, by constructing and releasing \textsc{SkillMatch}, a first-of-its-kind comprehensive benchmark for the skill relatedness task.
The dataset is available at \url{https://huggingface.co/datasets/jensjorisdecorte/SkillMatch-1K}.
It is based on a robust distillation of expert knowledge drawn from millions of job ads.
We start by describing the construction of the \textsc{SkillMatch} benchmark in \secref{sec:benchmark}.
We set out to evaluate commonly used representation methods for skill relatedness, listed in \secref{sec:methodology}, including a proposed new self-supervised objective to adapt a Sentence-BERT model~\cite{reimers2019sentence}.
This objective is specifically designed to enhance skill-skill relations based on co-occurrence in job ads.
Finally, we evaluate its performance of the new \textsc{SkillMatch} benchmark, and show the superior performance of our proposed self-supervised Sentence-BERT adaptation method in \secref{sec:experiments}.

%---------------------------------------------
\section{Construction of \textsc{SkillMatch}}
\label{sec:benchmark}
%=============================================
We create \textsc{SkillMatch}, a first-of-its-kind benchmark for the task of skill relatedness in the labor market. 
The creation of \textsc{SkillMatch} relies on expert knowledge embedded across millions of job ads.
More specifically, we observe that hiring managers explicitly mention their equivalent preference toward two or more related skills. 
One example of such an expression in a job ad is the following: \textit{``Experience with platforms such as \underline{Kubernetes}, \underline{KafKa}, or \underline{EKS}''}.
This lexical pattern, characterized by phrases like ``such as'', indicates that the underlined skills are considered equally relevant to the job requirements by the hiring manager.
When two skills are repeatedly mentioned together in this manner, it indicates that they are strongly related, and we base the creation of \textsc{SkillMatch} on this insight.
Through manual inspection, we construct two groups of lexical patterns, for a total of 15 patterns, that indicate these skill relations:

\begin{enumerate}
    \item \text{\textbf{X} \{and/or\} \{other/related/similar/equivalent\} ... \textbf{KW}} (8 patterns)
    \item \text{\textbf{KW} ... \{such as/including/especially/for example/e.g./i.e.\} \textbf{X}} (7 patterns)
\end{enumerate}

Here, \textbf{X} refers to the location in which related skills are expected to be enumerated.
Furthermore, to ensure relevancy of the sentence, the other side of the sentence should contain one keyword out of \textit{skill, technique, knowledge, experience, background}, at least within a 30 character range, as indicated by \textbf{KW}.
Stemming is used to increase the recall of this filter.
We note that the patterns are strongly inspired by the more commonly known Hearst patterns~\cite{hearst1992automatic}.
Nearly 32 million job ads were scanned for these patterns. 
All ads are written in English and posted in the United States in 2024.
A total of 360,449 unique sentences were found to adhere to these patterns.

The related skills are extracted from the sentence using a simple few-shot learning strategy, constructing a prompt with examples.
We use the recently introduced Gemini 1.5 Flash\footnote{\url{https://deepmind.google/technologies/gemini/flash/}} model for its strong performance and cost efficiency.
We refer to \appref{app:prompts} for details on the prompt.
Afterwards, the casing of all extracted skills is normalized towards its most frequent form, such that for example both ``css'' and ``CSS'' are represented by their most common variant ``CSS''.
Finally, related skill pairs are defined as those that co-occur at least 3 times, and have a conditional probability of 25\% or above, in each direction.
This dual criterion ensures that only closely related skills are considered as positive examples.
Negative pairs were generated by randomly selecting skills that never appeared together in the lexical patterns, maintaining an equal number of negative and positive pairs to balance the benchmark.
The complete benchmark contains 1,000 positive pairs and an equal amount of negative pairs.
Some examples of related skill pairs in \textsc{SkillMatch} are shown in \tabref{tab:examples}.

\begin{table*}[htbp]
\fontsize{9pt}{9pt}\selectfont
    \centering
    \begin{tabular}{l l c}
    \toprule
    \textbf{Skill 1} & \textbf{Skill 2} & \textbf{Frequency} \\
    \midrule
    HTML & CSS & 705 \\
    grammar & spelling & 137 \\
    deep learning&  natural language processing & 66 \\
    GDPR & CCPA & 41 \\
    AS9100 & ISO 9001 & 14 \\
    paid search & paid social & 7 \\
    Single Sign-On (SSO) & Multi-Factor Authentication (MFA) & 6 \\
    payroll software & benefits software & 5 \\
    \bottomrule
    \end{tabular}
    \caption{Examples of related skill pairs in \textsc{SkillMatch}.}
    \label{tab:examples}
\end{table*}

%---------------------------------------------
\section{Methodology}
\label{sec:methodology}
%=============================================

We use representation learning as the basis for expressing skill relatedness.
In this setup, a model takes a skill phrase as input and transforms it into a multidimensional vector.
The relatedness of a pair of skills is then indicated by the cosine similarity of their respective vectors.

\subsubsection{Static vector baselines} We implement static vector methods to represent and compare skills, as they have traditionally been used a lot to represent skills.
We implement both the Word2Vec Skip-gram~\cite{wordtovec} and the fastText~\cite{bojanowski-etal-2017-enriching} algorithms.
For both algorithms, we use an off-the-shelf model trained on generic text, and we train a domain-specific model on a large corpus of job ads.
When a skill consists of multiple words in the Word2Vec models, the weighted sum of their constituent words' vectors is used to represent a phrase, where the inverse frequency of the words in the training corpus are used as weights.

\subsubsection{Contextual models}

Static word vectors fail to represent the different meanings of words based on the context.
In contrast, transformer-based models such as Sentence-BERT produce contextual representations that capture the semantic meaning of the full input~\cite{reimers2019sentence}.
We propose a new method to adapt a pre-trained Sentence-BERT model for the task of skill relatedness, based on the co-occurrence of skills in a large corpus of job ads.
It requires a preprocessing step of extracting all skill spans from a job ad, which can be achieved through dedicated models as in \cite{zhang-etal-2022-skillspan} or through the use of an LLM.
This results in a corpus where each job ad is represented by a list of skill spans, in order of appearance.
A job ad with $N$ skill spans is then converted into $N-1$ positive training pairs, each consisting of two adjacent skills in the list.
These positive pairs are used to adapt the Sentence-BERT model with a contrastive learning objective, employing in-batch negatives through the InfoNCE loss function, as proposed by \cite{Henderson2017EfficientNL}.
This optimization leads the model to produce similar vector representations for skills that frequently occur in succession, thus encoding the nuanced relationships between skills more effectively.

%---------------------------------------------
\section{Experiments and Results}
\label{sec:experiments}
%=============================================

For the generic Word2Vec model, we use \textit{word2vec-google-news-300}\footnote{\url{https://code.google.com/archive/p/word2vec/}}, which has 300-dimensional vectors, trained on the Google News dataset.
For the generic fastText vectors, we use \textit{cc.en.300.bin}, a 300-dimensional vector model trained on Common Crawl and Wikipedia~\cite{grave2018learning}.
To train their domain-specific versions, we use a dataset of 755k job ads, also from 2024.
Note that these are not included in the dataset used to construct \textsc{SkillMatch}.
For the general domain Sentence-BERT model we use a model that is pre-trained on over 1B pairs of semantically similar sentences~\cite{reimers2019sentence}.\footnote{\url{https://huggingface.co/sentence-transformers/all-distilroberta-v1}}
We fine-tune this base model through contrastive learning on the adjacent skill phrase pairs.
The model is trained on a total of over 8.2 million skill phrase pairs.
We refer to \appref{app:details} for details on training hyperparameters.
Performance on \textsc{SkillMatch} is measured by the area under the precision-recall curve (AUC-PR) for the pooled positive and negative pairs in the benchmark, where the cosine similarity between the skill embeddings represents the extent to which two skills are related according to the learned representation model.
Additionally, for each skill in a given positive pair, the reciprocal rank of the matching skill is calculated, among the ranking of all other skills in the benchmark (from both positive and negative pairs). We report the mean reciprocal rank (MRR).

\subsection{Results}

The performance on \textsc{SkillMatch} for each method is shown in \tabref{tab:results}.
Our contrastive learning objective for the Sentence-BERT model significantly outperforms all other models.
This shows the effectiveness of modeling skill relatedness based on skill co-occurrence in job ads.
The domain-specific static vectors outperform their generic versions, despite being trained on fewer data.
Notably, the generic Word2Vec model performs much worse compared to the fastText model.
Further analysis reveals that this is due to words not being modeled in the generic Word2Vec model.
The number of skill phrases in \textsc{SkillMatch} without a Word2Vec representation drops from 193 to 36 when using the domain-specific model.
fastText does not suffer from this out-of-vocabulary issue by design.

\vspace{-0.3cm}

\begin{table*}[htbp]
\fontsize{9pt}{9pt}\selectfont
    \centering
    \begin{threeparttable}
    \begin{tabular}{l c c}
    \toprule
    \textbf{Method} & \textbf{AUC-PR} & \textbf{MRR} \\
    \midrule
    Word2Vec & 0.844 & 0.187\\
    Word2Vec\tnote{\textsc{ds}} & 0.922 & 0.300\\
    fastText & 0.913 & 0.262\\
    fastText\tnote{\textsc{ds}} & 0.941 & 0.325\\
    \midrule
    Sentence-BERT & 0.876 & 0.145 \\
    %Sentence-BERT_{\text{skill extraction}} & 0.867 & 0.114 \\
    Sentence-BERT\tnote{\textsc{ds}}\phantom{x}  & \textbf{0.969} & \textbf{0.357} \\
    \bottomrule
    \end{tabular}
    \begin{tablenotes}
    \small
    \item[\textsc{ds}] = domain-specific versions.
    \vspace{0.3cm}
    \end{tablenotes}
    
    \end{threeparttable}
    \caption{Comparison of model performance for the skill relatedness task. For the static vector models, the domain-specific versions refer to the models trained from scratch on job ads. For Sentence-BERT, the domain-specific model refers to its fine-tuned variant based on the proposed self-supervised skill co-occurrence objective.}
    \label{tab:results}
\end{table*}

\vspace{-0.9cm}

Despite its popularity, the pretrained Sentence-BERT model is outperformed by generic fastText vectors.
We assume this is partly because Sentence-BERT models are trained to represent full sentences, which might harm their ability to meaningfully represent shorter phrases.
As such, our fine-tuning strategy for Sentence-BERT effectively serves both purposes of learning domain-specific skill relations, as well as learning to effectively represent shorter phrases instead of full sentences.

\begin{figure}[ht]
\centering
\includegraphics[width=\textwidth]{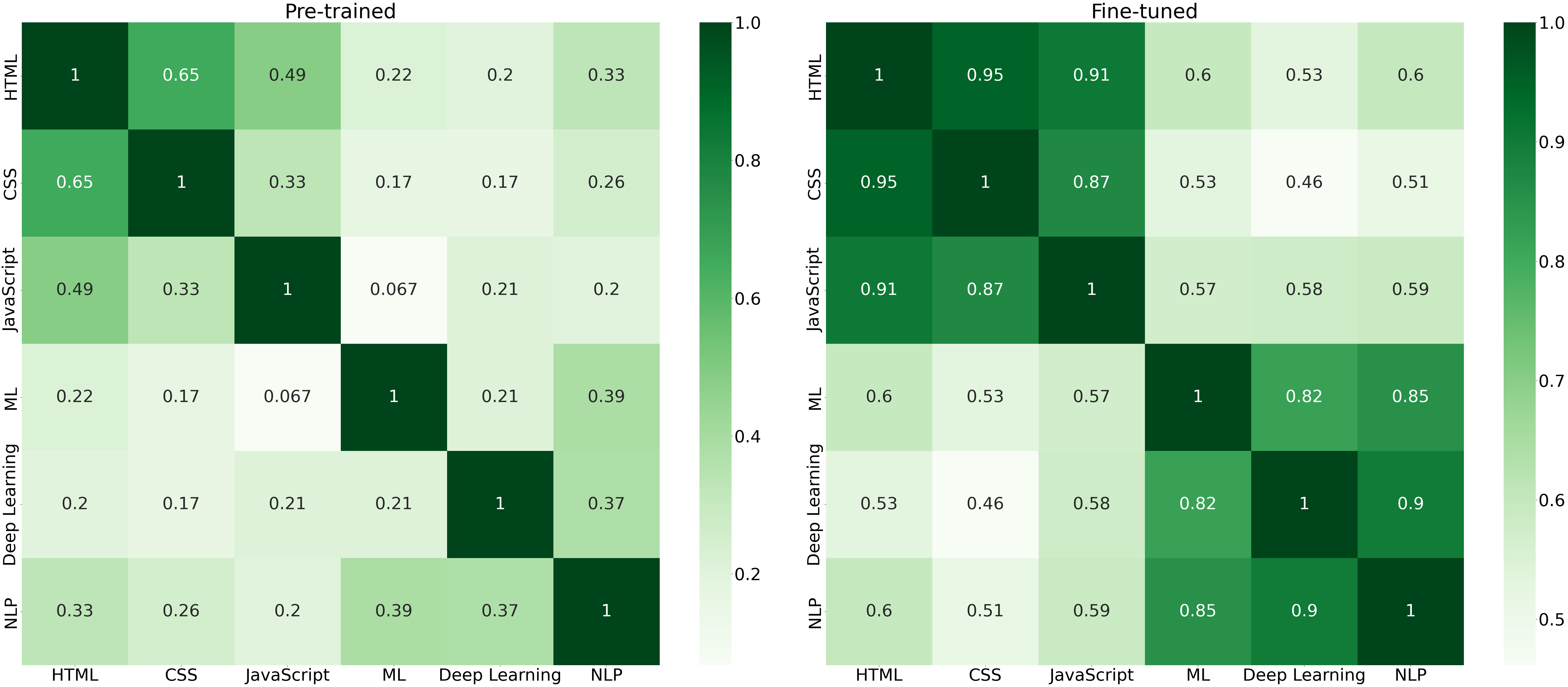}
\caption{A visualization of similarity scores between two clusters of skills (web development and machine learning), obtained through the pre-trained Sentence-BERT (left) and its fine-tuned counterpart (right). Clusters become clearly visible after fine-tuning.} \label{fig:heatmap}
\end{figure}

Finally, qualitative analysis of the skill relatedness scores allows inspecting the effect of our proposed Sentence-BERT fine-tuning approach.
\figref{fig:heatmap} displays one such example: a heatmap of similarity scores between two clusters of IT skills, being three web development related (\textit{HTML, CSS, JavaScript}) and three machine learning related skills (\textit{ML, Deep Learning, NLP}).

\section{Conclusion}
We introduce \textsc{SkillMatch}, a benchmark for evaluating skill relatedness, based on expert knowledge mined from millions of job ads.
We propose a new self-supervised technique to adapt a Sentence-BERT model for the task of skill relatedness, and show its effectiveness on \textsc{SkillMatch}.
With this contribution we hope to enable more research on the intrinsic evaluation of skill representation models, eventually to increase the accuracy and transparency of skill-based recommendation systems.

\section*{Limitations}

\textsc{SkillMatch} consists of binary annotations, while in reality, skill relatedness is a continuous spectrum, as some pairs of skills are more related than others.
It is not obvious how continuous values for relatedness could be accurately deduced from expert knowledge in job ads.
Secondly, applications that operate on a broader scale require an extension towards more languages and regions to ensure robust and accurate modeling of skill relations in different languages and cultures.
Finally, there are other interesting relations between skills, such as \textit{part-of} relations, that are not in the scope of this work.

\section*{Acknowledgments}

This project was funded by the Flemish Government, through Flanders Innovation \& Entrepreneurship (VLAIO, project HBC.2020.2893).

\bibliographystyle{splncs04}
\bibliography{references}

\begin{subappendices}
\renewcommand{\thesection}{\Alph{section}}

\newpage

\section{Prompts}
\label{app:prompts}

The following prompt is used to extract skill spans from job ads, as a preprocessing step for the Sentence-BERT fine-tuning procedure.
Note the placeholder \text{[Job Ad]}, which is replaced by the job ad from which skill spans need to be extracted.

\begin{tcolorbox}[fontupper=\footnotesize, colback=lightgray!20, boxrule=0.5pt, arc=4pt, boxsep=0pt, left=4pt, right=4pt, top=2pt, bottom=2pt, colframe=black, sharp corners]
\texttt{\underline{System instruction}\\
You are a helpful HR expert. Your mission is to list all relevant skills mentioned in job ad responsibilities and qualifications, as a comma-separated list. Don't mention vague terms like `relevant experience'.
\\
\\
\underline{Prompt}\\
\# Vacancy 1\\
As a Senior Software Engineer, you'll be a cornerstone of our Integrations team, developing and maintaining powerful APIs that enable seamless integrations. You'll be surrounded by brilliant minds eager to learn from your experience and insights. Your expertise will help scale our existing foundation and ensure we remain lean, smart and fun. If you're a collaborative communicator who thrives in dynamic environments and finds joy in mentoring colleagues, we want you on our team!\\
Job requirements\\
You have 5+ years of practical experience in a software engineering role, preferably at different company stages\\
You've gained deep expertise in Python, Django and/or Typescript backends\\
You've system design experience with a focus on scalability and performance\\
You're proficient in software development methodologies\\
You've demonstrated excellent mentorship and communication skills\\
Apply now!\\
\\
\# Skills 1\\
Software engineering, Python, Django, Typescript backends, system design, scalability, system performance, software development methodologies, mentorship, communication skills.
\\
\\
\# Vacancy 2
\\\text{[Job Ad]}
\\
\\
\# Skills 2}
\end{tcolorbox}

\newpage

The following prompt was used to extract the related skills from the sentences detected by the lexical patterns. Placeholder \text{[sentence]} is replaced by the sentence from which the related skills should be extracted.

\begin{tcolorbox}[fontupper=\footnotesize, colback=lightgray!20, boxrule=0.5pt, arc=4pt, boxsep=0pt, left=4pt, right=4pt, top=2pt, bottom=2pt, colframe=black, sharp corners]
\texttt{\underline{System instruction}\\
Given a sentence from a job ad, deduct whether it contains a list of equivalently desired skills. Do not include vague terms like 'experience' or 'knowledge'.
\\
\\
\underline{Prompt}\\
Sentence: Knowledge of supply chain management concepts such as product purchasing, pricing structures, and inventory control.\\
Equivalent skills list: product purchasing, pricing structures, inventory control\\
\\
Sentence: Comprehensive writing skills, including proper punctuation and grammar, organization, and formatting.\\
Equivalent skills list: punctuation, grammar, organization of texts, formatting documents\\
\\
Sentence: Knowledge of accounting operations to include all aspects such as accounts receivable accounts payable, etc.\\
Equivalent skills list: accounts receivable, accounts payable}
\\
\\Sentence: \text{[sentence]}
\\
Equivalent skills list:
\end{tcolorbox}

\newpage

\section{Training details}\label{app:details}

\subsection{Word2Vec}

For the domain-specific Word2Vec model, we use the skip-gram algorithm.
We use the popular gensim library to train the model~\cite{rehurek_lrec}.
We set the vector size to 300, which is the same as for the generic Word2Vec model.
Negative sampling is used with 5 negatives.
A minimum of 5 occurrences is set for words to be represented in the model.
The window size is set to 5.

\subsection{fastText}

For the domain-specific fastText model, we also use the skip-gram algorithm.
We use the fastText library in Python to train the model.\footnote{\url{https://fasttext.cc/}}
We set the vector size to 300, which is the same as for the generic fastText model.
Negative sampling is used with 5 negatives.
A minimum of 5 occurrences is set for words to be represented in the model.
The window size is set to 5.

\subsection{Sentence-BERT fine-tuning}

The contrastive training is implemented using the popular SBERT implementation in Python~\cite{reimers2019sentence}.\footnote{\url{https://www.sbert.net/}}
We keep the default value of 20 for the ``scale'' hyperparameter \textit{alpha}.
We train for 1 epoch.
The positive pairs are randomly shuffled into batches of 128 pairs.
We use the AdamW optimizer with a learning rate of 2e-5 and a ``WarmupLinear'' learning rate schedule with a warmup period of 5\% of the training data.
Automatic mixed precision (AMP) was used to speed up training.
Training was accelerated using an Nvidia V100 GPU.

\end{subappendices}

\end{document}